\documentclass[conference]{IEEEtran}
\IEEEoverridecommandlockouts
\usepackage{cite}
\usepackage{amsmath,amssymb,amsfonts}
\usepackage{algorithmic}
\usepackage{graphicx}
\usepackage{textcomp}
\usepackage{xcolor}
\usepackage{url}
\def\BibTeX{{\rm B\kern-.05em{\sc i\kern-.025em b}\kern-.08em
    T\kern-.1667em\lower.7ex\hbox{E}\kern-.125emX}}
\begin{document}

\title{Evaluating Cascaded Methods of Vision-Language Models for Zero-Shot Detection and Association of Hardhats for Increased Construction Safety}

\author{\IEEEauthorblockN{Lucas Choi}
\IEEEauthorblockA{\textit{Archbishop Mitty} \\
lucasleechoi@gmail.com}
\and
\IEEEauthorblockN{Ross Greer}
\IEEEauthorblockA{\textit{University of California Merced} \\
rossgreer@ucmerced.edu}
}


\maketitle

\IEEEpubidadjcol

\begin{abstract}
This paper evaluates the use of vision-language models (VLMs) for zero-shot detection and association of hardhats to enhance construction safety. Given the significant risk of head injuries in construction, proper enforcement of hardhat use is critical. We investigate the applicability of foundation models, specifically OWLv2, for detecting hardhats in real-world construction site images. Our contributions include the creation of a new benchmark dataset, Hardhat Safety Detection Dataset, by filtering and combining existing datasets and the development of a cascaded detection approach. Experimental results on 5,210 images demonstrate that the OWLv2 model achieves an average precision of 0.6493 for hardhat detection. We further analyze the limitations and potential improvements for real-world applications, highlighting the strengths and weaknesses of current foundation models in safety perception domains.
\end{abstract}


\section{Introduction}
The use of hard hats in construction is an instance where appropriate safety covering can prevent injury or death and may be enhanced by IoT-style worksite safety monitoring. Annually, there are more than 800 casualties of construction workers as a result of workplace accidents \cite{OSHA-transcript}. In 2020, the private construction industry reported 1,008 deaths, marking the highest number of fatalities among all private industries \cite{U.S.-Bureau-of-Labor-Statistics}. Many of these fatal injuries can be lessened or avoided with proper wearing of hard hats.

Hard hats provide essential protection against head injuries caused by falling objects, electrical hazards, and collisions, which are prevalent risks in construction and industrial settings. Given the increased risk of fatal injuries among construction workers, it is crucial to enforce stringent regulatory safety measures within the workplace. 

OSHA's Section 1926.100 requires protective helmets for employees in areas with possible head injury risks \cite{Occupational-Safety-and-Health-Administration}. Despite regulations mandating hard hat use, compliance is inconsistent, leading to preventable injuries.
Having a guideline supported by camera detection may help workers be more aware of when and where to wear hard hats. By properly detecting workers' hard hat status through visual surveillance, sites can raise awareness of the issue and assist with the enforcement of hard hat-wearing.

The question we explore in this research is to what degree foundation model approaches are ready for use with real-world data in these safety perception domains and where their strengths and weaknesses may lie. Therefore, in this research, we make research contributions of (1) the evaluation of zero-shot methods using foundation vision-language models (VLMs) of hard-hat detection and association and (2) the creation of an evaluation benchmark dataset for hard hat detection, titled Hardhat Safety Detection Dataset, combining and filtering annotations of the following image source datasets: Hard hat workers dataset\cite{hard-hat-workers_dataset} and the SHEL5k Dataset\cite{otgonbold2022shel5k}. 

We begin by presenting a framework for application of the zero-shot detection method in our hardhat detection process using a cascaded detection method. Our experimental results with 5,210 images show that the OWLv2 model achieves an average precision of 0.6493 on hardhat detection. We also conduct a case study on several failed detections to analyze the limitations of using OWLv2 for hardhat detection, and discuss methods of improving the model for real-world robust detection. 

\section{Related Research}

Traditional machine learning object detection algorithms historically rely on manual annotations and specialized algorithms, which can be time-consuming and resource-intensive, especially as the models are limited to learning from provided datasets. Moreover, these methods often lack the flexibility to adapt to new environments or variations in hardhat designs \cite{Object-Detection-Review}. 

\begin{figure}
    \centering
\includegraphics[width=.48\textwidth]{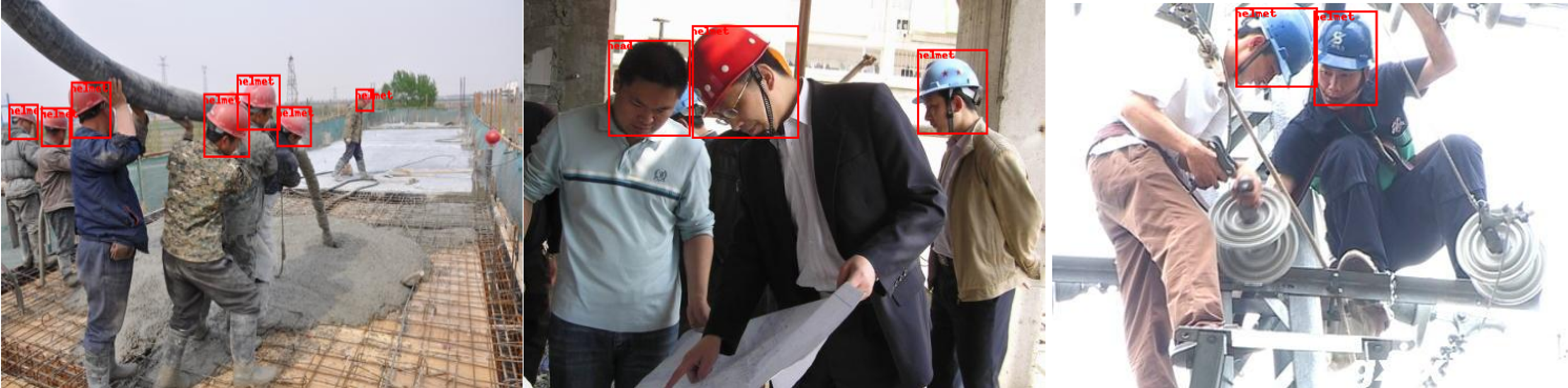} 
    \caption{Example images with ground truth bounding boxes from the Hard Hat Workers Dataset demonstrating the absence of person class annotations}
    \label{fig:faultannot}
\end{figure}

In contrast, foundation vision-language models, or VLMs, with their ability to generalize from text descriptions and visual features, offer a more adaptable solution. VLMs are a type of artificial intelligence that integrates both visual and textual data to perform tasks requiring a deep understanding of these two modalities, and have been useful in a variety of real-world safety applications \cite{greer2024towards, ghita2024activeanno3d, greer2024driver, gopalkrishnanmulti, greer2024language}. These models are designed to analyze images and interpret text, enabling them to carry out tasks such as image captioning, visual question answering, and multimodal reasoning.

Due to being pre-trained on a much larger magnitude of data, these models can approach unseen tasks with higher accuracy. Our research builds on the Vision Transformer for Open-World Localization (OWL-ViT) and the OWLv2 family of models found in \cite{minderer2022simple} and \cite{minderer2023scaling}, and is further detailed in \cite{choi2024evaluating}. 



We propose to evaluate foundational VLMs' zero-shot learning capability for their readiness for use in the detection of hard hats within real-world data. Other researchers have applied various methods towards the same task of hard hat detection. Xie et al. \cite{xie2018convolutional} proposed the CAHD algorithm, a convolutional neural network based hard-hat detection algorithm. This algorithm achieved a mAP of 54.6\% on the ImageNet Dataset\cite{deng2009imagenet}. However, the ImageNet dataset only shows objects in central focus, unlike real-world images where the object can be in the background, which we choose to address.

\section{Methodology}
Accurately detecting and enforcing the use of protective gear like hard hats is a task that remains challenging despite advancements in object detection models. The inconsistencies and incomplete annotations in existing datasets, such as the Hard Hat Workers Dataset \cite{hard-hat-workers_dataset}, further complicate this task, leading to unreliable performance metrics and hindering the development of effective safety solutions. Our methodology aims to address these challenges by releasing a new dataset that addresses the dataset issues and detailing our cascaded detection strategy using the OWLv2 model.

\subsection{Data Preprocessing}

We used the dataset provided by Hard Hat Workers Dataset\cite{hard-hat-workers_dataset} and SHEL5k Dataset\cite{otgonbold2022shel5k}.
The Hard Hat Workers Dataset gives three classes: helmet, head, and person.

There are 7,063 images and 5,000 images within the Hard Hat Workers Dataset\cite{hard-hat-workers_dataset} and the SHEL5k Dataset\cite{otgonbold2022shel5k}, respectively. However, the Hard Hat Workers Dataset is largely inconsistent with their annotations, as many of the images don't have any `person' class even though a person is evidently within the image as shown in Figure \ref{fig:faultannot}. There are a few exception cases where the annotation is labeled correctly. With these inconsistencies, it is impossible to get a representative metric of the accuracy of the OWLv2 model, as the detections will be judged inaccurate for nonexistent annotations. Therefore, we filtered out all the images from this dataset and separated only the correctly labeled images with the person class.

\begin{figure}
    \centering
\includegraphics[width=.07\textwidth, height=2.3cm]{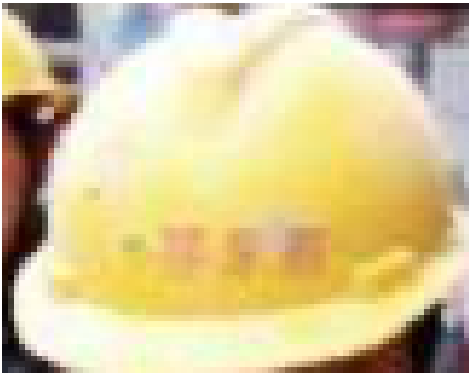}
\includegraphics[width=.07\textwidth, height=2.3cm]{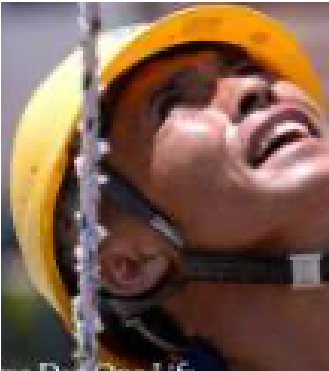}
\includegraphics[width=.07\textwidth, height=2.3cm]{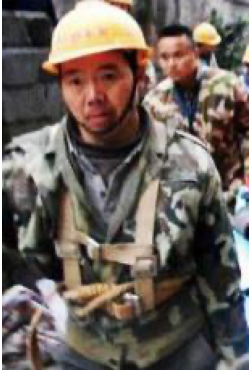}
\includegraphics[width=.07\textwidth, height=2.3cm]{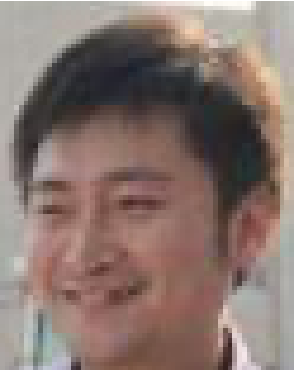}
\includegraphics[width=.07\textwidth, height=2.3cm]{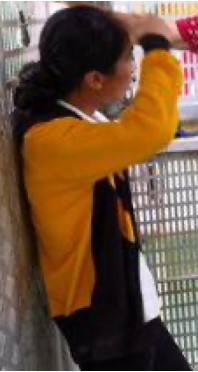}
\includegraphics[width=.07\textwidth, height=2.3cm]{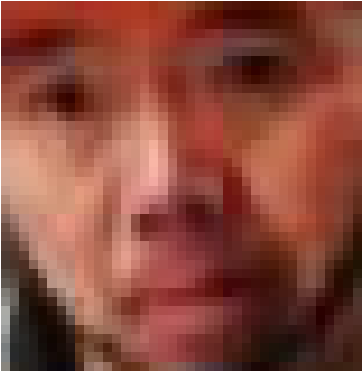}

    \caption{Example instances of classes cropped from the SHEL5k dataset. From left to right: Helmet, Head with Helmet, Person with Helmet, Head (Head without Helmet), Person without Helmet, and Face. However, not every object in the SHELF5k dataset receives every annotation it belongs to.}
    \label{fig:exampleinstances}
\end{figure}

Additionally, we selectively chose the annotations from the SHEL5k dataset to align with the Hard Hat Workers dataset to maintain consistency. The SHEL5k originally consisted of the following 6 classes: helmet, head with helmet, person with helmet, head (head without helmet), person without helmet, and face. We show sample data of these classes in Figure \ref{fig:exampleinstances}.

This resulted in a total of 5,210 images from both datasets for our benchmark dataset, titled Hardhat Safety Detection Dataset \footnote{https://www.kaggle.com/datasets/lcsc00/hardhatdetect}.

The new dataset contains 16,652 Helmets (Head with helmets), 20,631 Persons, and 6,158 Heads for our cascaded approach and 19,856 Helmets for our nested and direct approach. The frequencies are shown in Figure \ref{tab:GroundTruth}

\begin{table}[]
    \centering
    \caption{Ground Truth Data}
    \begin{tabular}{c|c}
        
        Class & Frequency \\
        \hline \hline
        Head with Helmets & 16,652 \\
        \hline
        Helmets & 19,856 \\
        \hline 
        Heads & 6,158 \\
        \hline 
        Persons & 20,631\\
        \hline  \hline
        Total & 63,297\\
        
    \end{tabular}
    \label{tab:GroundTruth}
\end{table}



\subsection{Cascaded Approach}

We implemented a cascaded detection approach with OWLv2 to determine whether construction workers are wearing hard hats. This method starts to detect a higher-level class, ‘person’ and then progressively narrows down the image to identify sub-classes in the sequence of ‘head’ and ‘hard hat’. The benefit of such an approach is an automatic association of a higher-level class with its lower-level attributes or features \cite{gopalkrishnan2023robust, greer2024patterns}.

Our approach works by recognizing that categories like `person' and `head' are related in a hierarchy. First, we detect the broader category, `person'. Once we identify a person in the image, we then focus on the `head' within that area. From there, we can further determine whether the person is wearing a hardhat. This allows the association of hard hat wearing to the person while improving the detection rate by concentrating on specific parts of the image

We begin with our person detection. We give an image and the text prompt, `person', as input to OWLv2. The image is normalized and resized while the text is encoded by CLIPTokenizer, from \cite{radford2021learning}, to be wrapped by the processor with the normalized image. 

After detecting a person, we extract and rescale the corresponding bounding boxes, cropping the relevant section from the image. This cropped image is used as input for the second step, where OWLv2 is prompted with the text `head' to detect the person's head.

\begin{figure}
    \centering
    \includegraphics[width=0.5\textwidth]{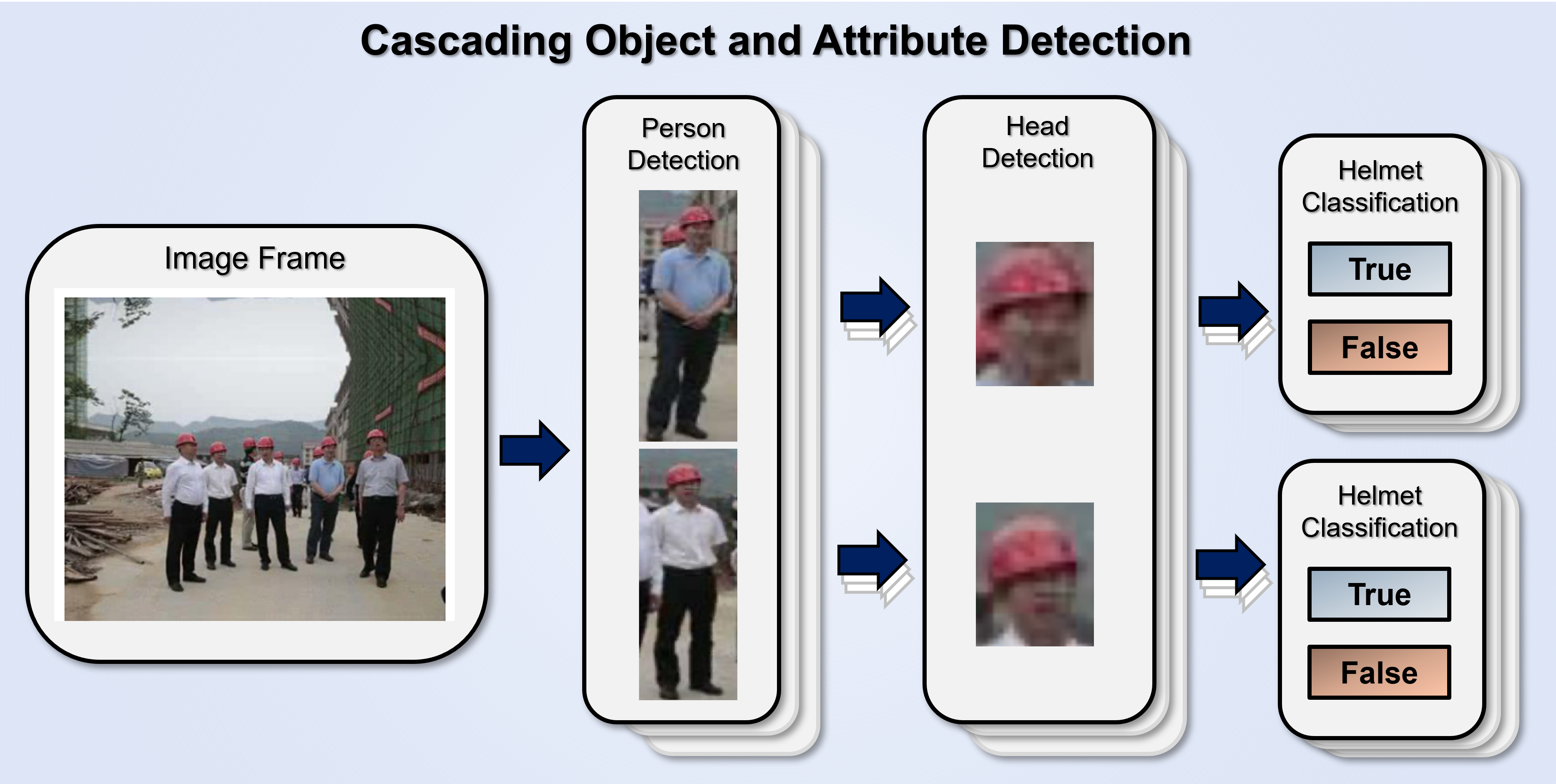}
    \caption{Diagram of Cascaded Object and Attribute Detection. From the original image, we detect all instances of persons. Within each person, we detect a head and then detect a helmet within the head. If a helmet detection is made, we classify the head as helmet-wearing. All detections, including helmet detection for the purpose of classification, are performed using OWLv2.}
    \label{fig:HardHat-Process}
\end{figure}

\begin{figure}
    \centering
    \includegraphics[width=0.5\textwidth]{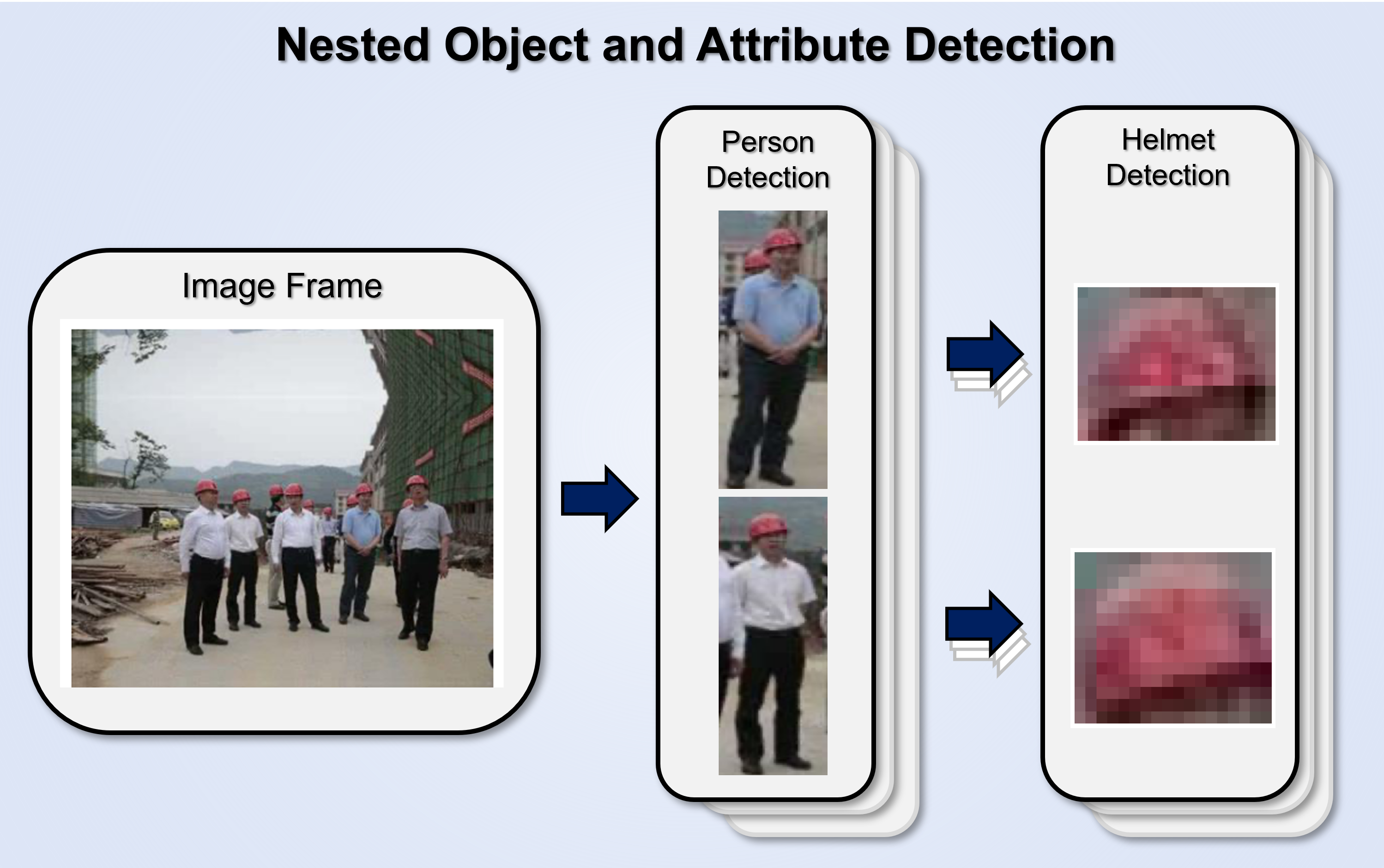}
    \caption{Diagram of Nested Object and Attribute Detection. Diagram of Cascaded Object and Attribute Detection. From the original image, we detect all instances of persons. Within each person, we detect a helmet. All detections are performed using OWLv2.}
    \label{fig:NProcess}
\end{figure}

Finally, we proceed to the last level of detection. The bounding box from the head detection is again cropped from the original image and used as input, along with the text prompt `helmet'. At this stage, we classify whether a hard hat is present, determining a boolean value based on the detection. This process does not require finding another bounding box but rather identifies the presence or absence of the hard hat as an attribute of the detected head. This cascaded detection process is shown in Figure \ref{fig:HardHat-Process}, and the nested is in Figure \ref{fig:NProcess}. 


\section{Experimental Method and Evaluation}

Using the cascaded object and attribution detection algorithm detailed in the previous section, we performed hard hat detection on the combined dataset, with further implementation details described in this section. The evaluation metric we utilize for our task is the average precision (AP), as well as precision and recall across thresholds. 

We performed detections within the 5,210 images in the combined dataset.
Our detections were conducted across several thresholds, ranging from 0.05 to 0.5, to calculate the precision-recall curve. With this threshold sweep, we calculated the AP.

The threshold for OWLv2 is the minimum confidence threshold to use to filter out predicted boxes. It is the value that determines the minimum level of confidence required for OWLv2's detection to be accepted as a valid positive detection. OWLv2 calculates confidence through logits per image.

The intersection over union (IoU) is defined as the ratio of the area of overlap between the predicted bounding box and the ground truth bounding box to the area of their union. We use this as our evaluation metric by observing if the IoU is greater or less than 0.5 to determine whether detection is a true positive or false positive. The IoU is calculated by: \begin{math}
IoU=\frac{A\cap B}{A\cup B}
\end{math} or \begin{math}
IoU=\frac{Area\:of\:Overlap}{Area\:of\:Union}
\end{math}, where \begin{math}A\end{math} stands for the predicted bounding box and \begin{math}B\end{math} stands for the ground truth bounding box.
 
To evaluate our cascaded methodology of a hierarchical multi-stage detection and observe its effectiveness, we additionally tested a nested detection approach and a direct detection approach of hard hats. The nested detection was performed by detecting persons and then detecting hard hats within the person's bounding box. However, we note that the direct detection approach does not address the association task of 
the hard hat to a person, detecting hard hats on the ground. As with the complete cascaded detection, we performed a hyperparameter sweep of the threshold for both processes to compare the APs of hard hat detection.

Additionally, for our data, when using the cascaded approach, we treated all people (whether wearing a helmet or not) as one single class and did not focus on detecting annotations of faces or isolated helmets since we only concerned ourselves with helmets associated with a person's head. However, when testing nested detection and direct detection, as described in Section IV, we used the isolated helmet class rather than the head with helmets.

The definition of ground truth varies for different categories. In person detection, all instances of persons are considered as ground truth, and for head prediction, every head without a helmet is the ground truth. However, for helmets, we include all helmets, even those on the ground, when computing the metrics. This causes errors in false negatives for the nested detection approach as it can only detect helmets on a person. Re-annotating the dataset is needed to resolve this issue.

\subsection{Results}

\begin{figure}
    \centering
    \includegraphics[width=.4\textwidth]{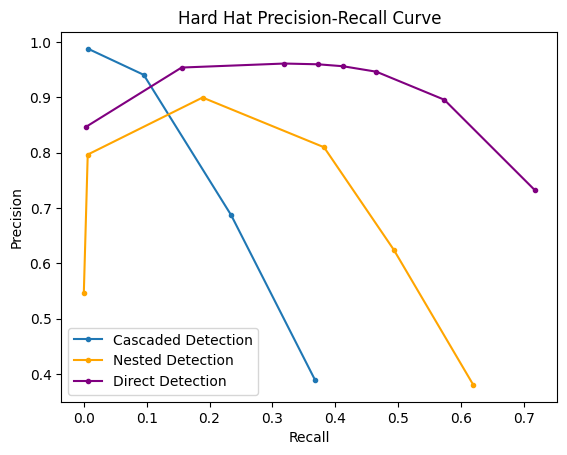}
    \caption{Precision-Recall Curves of the Different Approaches to Hard Hat Detection. The Direct Detection has a strong performance, maintaining a high precision throughout all recall values. The Nested Detection contrasts by starting with at a lower point, peaking at 0.8995, and then quickly decreasing. Finally, the Cascaded Detection demonstrates the poorest performance, starting off with high precision but rapidly declining, making its average precision the worst.}
    \label{fig:HardhatPRs}
\end{figure}

We analyzed the detection accuracies separately for hard hats, heads, and persons. The results of the evaluation of the effectiveness and efficiency of each detection strategy for hard hat detection are provided in Table \ref{tab:strategies}. We notice that as we remove the layers of detection, the precision and recall improve significantly, as shown in Figure \ref{fig:HardhatPRs}.

\begin{figure}
    \centering
    \includegraphics[width=.4\textwidth]{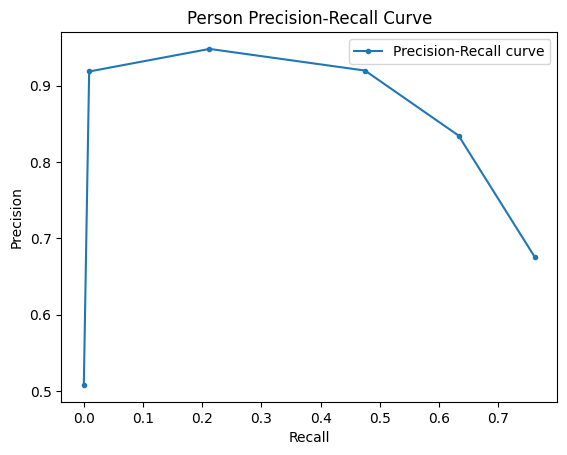}
    \caption{Precision-Recall Curve of Person Detection. At higher thresholds, OWLv2 is not able to make enough relevant detections. Throughout lower thresholds, the curve decreases slowly, suggesting high performance in person detection.}
    \label{fig:PRPerson}
\end{figure}

 The results of the person detection for both the nested and the cascaded detection approach are shown in Figure \ref{fig:PRPerson}, with an average precision, calculated by area under the curve, of 0.6767. However, we notice an abnormal low point at the beginning of the precision-recall curve, referencing that as the threshold was raised, the OWLv2 made too few detections for any relevant detections to be made.

\begin{figure}
    \centering
    \includegraphics[width=.4\textwidth]{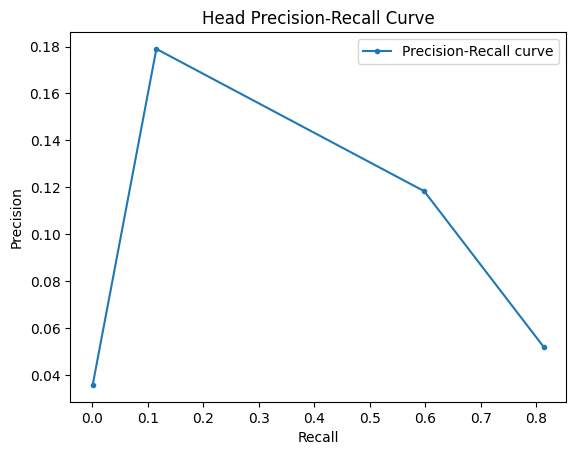}
    \caption{Precision-Recall Curve of Head Detection. At higher thresholds, OWLv2 can hardly make any relevant detections. The curve peaks at 0.179 and decreases. This demonstrates overall poor performance in head detection, having low precision, especially with higher recall values.}
    \label{fig:PRHead}
\end{figure}

The accuracy of the head class (heads without helmets) is shown in Figure \ref{fig:PRHead} with an average precision of 0.1024.

\begin{table}[]
    \centering
    \caption{Comparative evaluation of detection methods through Average Precision(AP) of Hard Hat detection}
    \begin{tabular}{c|c}
        Detection Method & AP (Area under the curve)\\
        \hline \hline
        Direct (Hard Hats) & 0.6493 \\
        \hline
        Nested (Person $\rightarrow$ Hard Hats) & 0.4672 \\
        \hline
        Multistage (Person $\rightarrow$ Head $\rightarrow$ Hard Hats) & 0.2699\\
        
    \end{tabular}
    \label{tab:strategies}
\end{table}

\section{Discussion}
As shown in Table \ref{tab:strategies} and Figure \ref{fig:HardhatPRs}, the removal of cascading levels within the detection strategy improves performance. This is due to the fact that as we continue to crop the image multiple times, the image quality gets progressively worse, meaning with more stages of detection and cropping, the more inaccurate the final stage of detection will be. Additionally, the final stage, the hard hat detection, depends on the upper levels of detection, meaning that if the head or person is detected inaccurately or not detected at all, the hard hat has 0 chance of being detected, further reducing accuracy.

Therefore, adding layers to the detection approach adds more variables of error and reduces accuracy, proving that a direct detection approach is the most accurate. However, to address the task of associating a hardhat with a person, having at least the nested detection approach is necessary.

Additionally, the head detection within the cascaded detection was shown to have very poor performance. This may extend the problem described above. As helmet detection doesn't have the best accuracies due to the reduction of quality by cropping, the rest of the detected heads are classified as heads without helmets. This means that there are many false positives for heads without helmets and many false negatives for helmets, resulting in a decrease in the precision of the head detection and the recall of the helmets.

\begin{figure}

  \centering
\includegraphics[width=.4\textwidth]{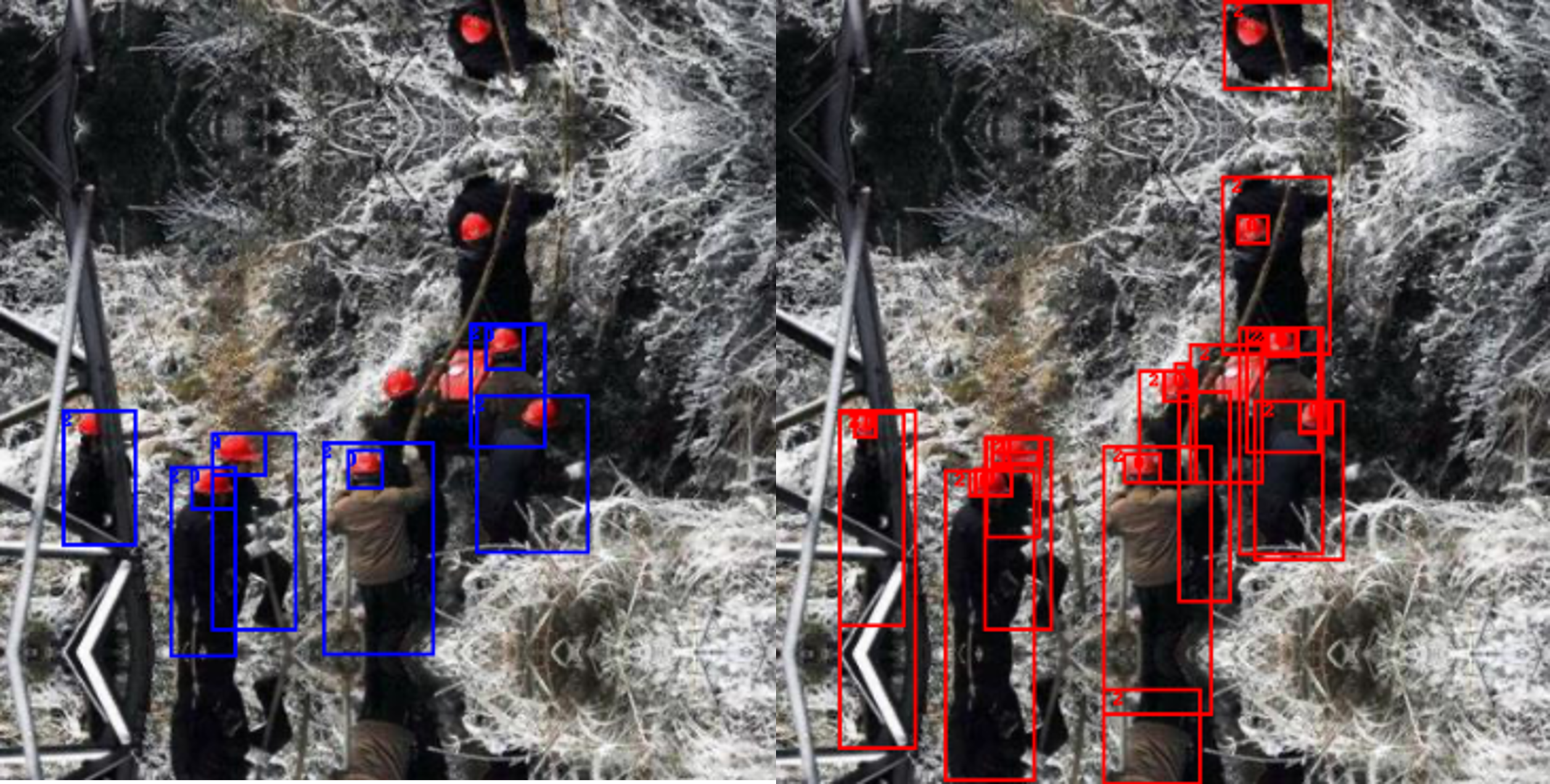}
\includegraphics[width=.4\textwidth]{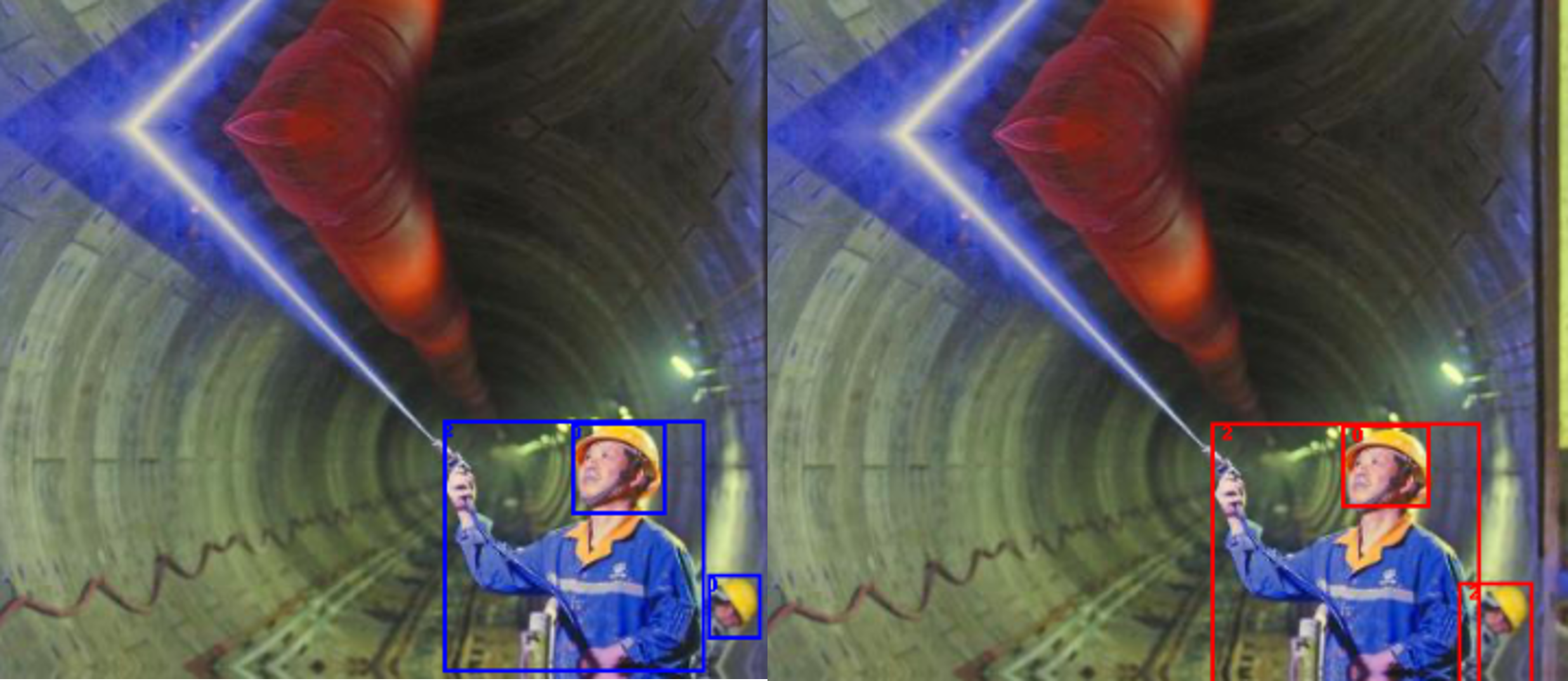}
\includegraphics[width=.4\textwidth]{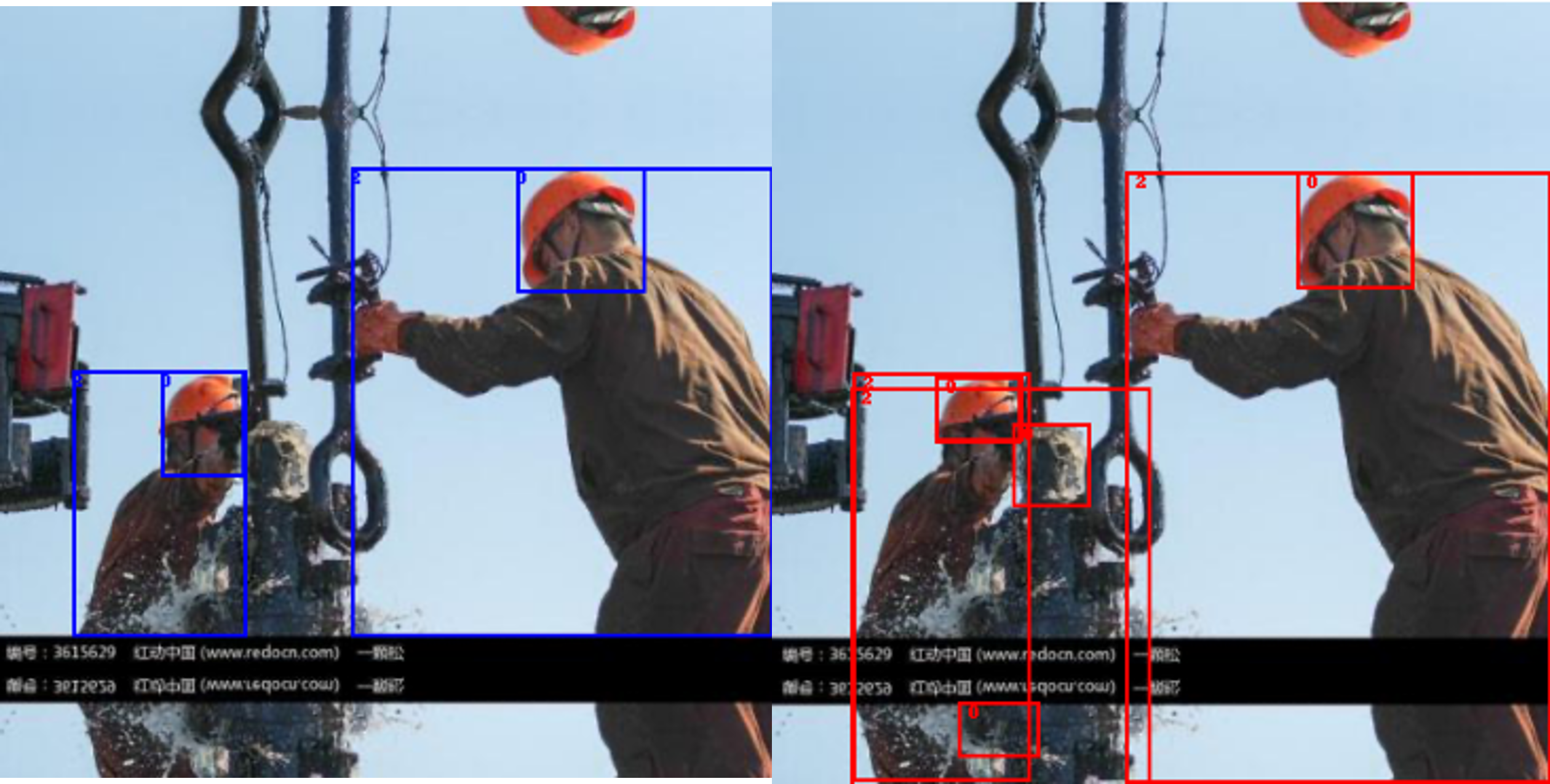}
\caption{Sample images of the dataset of different angles with different environments, showing potential sources of errors. The left image is the ground truth of the annotations of the cascading approach, and the right is the prediction. From top to bottom: incomplete annotations, obstruction, similar-looking objects. As shown in the top row, OWLv2 on the right detects the few people at the top of the image. However, there are no ground truth bounding boxes for those persons, as shown in the left image. Additionally, the helmet(head with helmet) class follows this trend, only having 4 instances when it should have 9. This suggests that some failures may be due to inaccuracies in the ground truth annotations, as similar issues have been observed in other cases. Furthermore, in the middle row, the construction worker in the background is obscured by the person in front, and their hunched posture makes it difficult to recognize the shape of their head and helmet accurately. OWLv2 detects the person but not the helmet, while the ground truth only provides the helmet of the individual in the back. Finally, in the third row, the model mistakenly identifies construction machinery as a person and erroneously classifies parts of the machinery as a helmet. This case highlights that OWLv2's performance may still fall short of expectations. }
\label{fig:samp}
\end{figure}

One nuance of removing the head detection step is that people who are holding helmets will be classified as helmet-wearing, even though they are not wearing it. When observing the data, we notice that possible sources of error include incomplete annotations, obstruction, and similar-looking objects, as shown in Figure \ref{fig:samp}. To mitigate these issues, capturing and analyzing multiple images of the same scene while improving annotations could help reduce errors and improve accuracy.

\section{Concluding Remarks and Future Research}

We have demonstrated the potential of using foundation vision-language models for zero-shot detection of hard hats to enhance safety on construction sites. By creating the Hardhat Safety Detection Dataset and using the OWLv2 model in a cascaded detection approach, we found that direct detection of hard hats yields higher accuracy compared to multi-stage cascaded methods due to image quality degradation and compounding errors. Despite these challenges, associating detected hard hats with individuals is crucial for practical safety, emphasizing the need for robust annotations in datasets.

Future research should focus on several key areas to enhance the efficacy and reliability of hard hat detection. Improving and expanding datasets with comprehensive annotations is essential to address missing instances and to develop separate annotations for helmet-person association and detection. Additionally, developing techniques to avoid quality reduction in the cascaded detection approach and exploring state-of-the-art models and hybrid approaches can offer significant improvements. Furthermore, reducing false positives and negatives remains a critical challenge, and techniques such as multi-frame analysis, context-aware detection, and incorporating additional sensory data should be explored to enhance detection reliability.

Future research can significantly improve occupational safety by addressing these areas. The advancement of VLMs holds great promise for creating safer work environments.


\bibliographystyle{ieeetr}
\bibliography{refs}
\end{document}